\documentclass{SCIS2021}

\usepackage{times}
\usepackage{latexsym}
\usepackage[T1]{fontenc}
\usepackage[utf8]{inputenc}
\usepackage{microtype}

\usepackage{graphicx}
\usepackage{booktabs}
\usepackage{amssymb}
\usepackage{amsmath}
\usepackage{color}
\usepackage{makecell}
\usepackage{appendix}
\usepackage{enumitem}
\usepackage[sort]{natbib}
\usepackage{wrapfig}
\usepackage{multirow}

\newcommand{\metricname}{LInCo} 
\newcommand{\metricsymbol}{\text{LInCo}}

\newcommand{\tabincell}[2]{\begin{tabular}{@{}#1@{}}#2\end{tabular}}

\begin{document}
\ArticleType{RESEARCH PAPER}
\Year{2020}
\Month{}
\Vol{}
\No{}
\DOI{}
\ArtNo{}
\ReceiveDate{}
\ReviseDate{}
\AcceptDate{}
\OnlineDate{}


%
%

\title{Equality before the Law: Legal Judgment Consistency Analysis for Fairness}

\author[1]{Yuzhong Wang}{{wyzthu@gmail.com}}
\author[1]{Chaojun Xiao}{{xcjthu@gmail.com}}
\author[1]{Shirong Ma}{{msr17.2000@gmail.com}}
\author[1]{Haoxi Zhong}{{zhonghaoxi@yeah.net}}
\author[1]{\\Cunchao Tu}{{tucunchao@gmail.com}}
\author[2]{Tianyang Zhang}{{zty@powerlaw.ai}}
\author[1]{Zhiyuan Liu}{{lzy@tsinghua.edu.cn}}
\author[1]{Maosong Sun}{{sms@tsinghua.edu.cn}}

\AuthorMark{Yuzhong Wang}

\AuthorCitation{Yuzhong Wang, Chaojun Xiao, Shirong Ma, et al}


\address[1]{Department of Computer Science and Technology Institute for Artificial Intelligence, Tsinghua University, Beijing, China}
\address[2]{Beijing Powerlaw Intelligent Technology Co., Ltd., China}

\abstract{
In a legal system, judgment consistency is regarded as one of the most important manifestations of fairness.
However, due to the complexity of factual elements that impact sentencing in real-world scenarios, few works have been done on quantitatively measuring judgment consistency towards real-world data.
In this paper, we propose an evaluation metric for judgment inconsistency, \textbf{L}egal \textbf{In}consistency \textbf{Co}efficient (\textbf{\metricname{}}), which aims to evaluate inconsistency between data groups divided by specific features (e.g., gender, region, race).
We propose to simulate judges from different groups with legal judgment prediction (LJP) models and measure the judicial inconsistency with the disagreement of the judgment results given by LJP models trained on different groups.
Experimental results on the synthetic data verify the effectiveness of \metricname{}.
We further employ \metricname{} to explore the inconsistency in real cases and come to the following observations: 
(1) Both regional and gender inconsistency exist in the legal system, but gender inconsistency is much less than regional inconsistency; 
(2) The level of regional inconsistency varies little across different time periods; 
(3) In general, judicial inconsistency is negatively correlated with the severity of the criminal charges.
Besides, we use \metricname{} to evaluate the performance of several de-bias methods, such as adversarial learning, and find that these mechanisms can effectively help LJP models to avoid suffering from data bias.
}


\maketitle

\section{Introduction}


Legal judgment consistency indicates the degree to which similar cases yields similar court decision in realistic legal systems. It is an essential component of judicial fairness, whether in the Civil Law system or the Common Law system~\citep{flynn2013making}. 
However, even though we have a well-developed legal system, cases are not always judged consistently in reality. Table~\ref{table:sample} gives an example where two cases are judged inconsistently. The two criminals behaved similarly, but they received quite different verdicts.
The inconsistency in legal judgment may be caused by many complex factors, like regional differences, public opinion, or judges' personal subjective thoughts. 
Different judgments to similar cases violate the principle of equality before the law, undermines the parties' rights and interests. Therefore, it is essential to develop efficient and effective methods to evaluate legal judgment inconsistency in the real world.

\begin{table}[ht]
\centering
\small
\caption{An example of inconsistency between two cases from different regions. In these two cases, the criminals behaved similarly, and the stolen items were of equal value, but they received different verdicts.}
\label{table:sample}
\begin{tabular*}{\textwidth} {p{0.975\columnwidth}}
\toprule
\textbf{Case A:} In a snack bar in Region A, Alice stole the victim's iPhone. After identification, the value of the stolen phone was RMB 4000. \\
\textbf{Imprisonment:} 6 months.~~~\textbf{Fine:} RMB 1000.\\\hline
\textbf{Case B:} When working in a restaurant in Region G, Bob found that the victim who was eating had hung his jacket on a stool, and a Samsung mobile phone (valued at RMB 4000) was in his jacket pocket, so he stole the cell phone while no one was watching. \\
\textbf{Imprisonment:} 9 months.~~~\textbf{Fine:} RMB 4000. \\
\bottomrule
\end{tabular*}
\end{table}

Legal inconsistency analysis has been studied for decades. \citet{anderlini2020legal} use a mathematical approach to analyze and compare judicial consistency under different legal systems in theory. And other approaches focus on analysing real-world data from a case-by-case perspective~\citep{reamer2005ethical,li2014a,edgely2009common,anderlini2014why}. However, few works focus on macro-consistency analysis towards real-world data, which will be important complementarity and support to qualitative research.


In this paper, we explore the quantitative analysis of judgment consistency. 
Judicial fairness requires similar cases to be judged similarly. But it is difficult to decide whether two cases are similar in real-world~\citep{xiao2019cail2019-scm:,zhong2020does}, which makes it challenging to analyze legal inconsistency by comparing the judgment results between similar cases. Therefore, we propose \textbf{L}egal \textbf{In}consistency \textbf{Co}efficient, \metricname{}, which focuses on the judicial inconsistency between groups divided by specific features (e.g., gender, region, race). We simulate judges from different groups with legal judgment prediction (LJP) models, which can predict the term of penalty from cases' fact description. 
Since LJP models are able to capture the characteristics within datasets and reflect the bias from the historical judgment in prediction~\citep{grgichlaca2018human}, we can train multiple LJP models as ``virtual judges'' on each data group, and the legal inconsistency coefficient is defined as their disagreement on the same cases.
Specifically, given several groups of cases divided by specific features, such as regions or gender, calculating \metricname{} includes two steps: 
(1) Train an LJP model as a virtual judge for each group and predict the term of penalty for all cases in test sets;
(2) Calculate \metricname{} as the average disagreement between different virtual judges on the test cases.


In experiments, we choose the Chinese legal system for analysis. China is a country with a large population and a vast territory, and there exist many complex factors (e.g., race, region) that cause inconsistency. Therefore exploring judicial inconsistency in the Chinese legal system is an urgent need.
Moreover, the Chinese government has published a large-scale dataset of formatted legal documents, which provides data support for our research. 

Our contributions are threefold:

(1) We propose \metricname{} to measure the legal inconsistency in real-world data. The results in the simulation study verify the effectiveness of \metricname{}, and show that the correlation coefficient between \metricname{} and the inconsistency factor $\beta$ is high ($\ge 0.94$), which proves the reliability of \metricname{}. (Section~\ref{section:experiment1})

(2) We conduct a series of region-specific (cases are judged in different regions) and gender-specific (defendants have different genders) experiments on real-world datasets. The results show that there exists inconsistency between different genders or different provincial-level administrative regions in the Chinese legal system. However, gender inconsistency is much less than regional inconsistency. We also discover that, in general, judicial inconsistency is negatively correlated with the severity of the criminal charges. (Section~\ref{section:experiment2})


(3) We use \metricname{} to evaluate the performance of several de-biasing methods, including adversarial learning. The results show that the methods are effective in decreasing \metricname{}, which indicates that the methods can mitigate the data bias for existing LJP models. (Section~\ref{section:experiment3})



We will release the datasets and source code once accepted to help the researchers make improvements on legal consistency analysis.

\section{Related Work}

\subsection{Legal Judgment Consistency}

Judicial inconsistency is a serious threat to the authority of the legal system and has attracted the attention of many researchers in the legal field. Most of them focus on the importance of consistency, causes of inconsistency, solutions, etc., from a case-by-case or theoretical perspective~\citep{reamer2005ethical,li2014a,edgely2009common,anderlini2014why,anderlini2020legal}. 
Meanwhile, many efforts have been devoted to developing fairness-aware machine learning algorithms~\citep{zafar2017fairnessa,dwork2012fairness,speicher2018a,berk2018fairness}, which focus on designing models to produce fair outcomes rather than analyzing the unfairness in existing data or legal systems. 
There are also some existing inconsistency coefficients~\citep{li2012inconsistency,jezewski2003analysis,korenius2004stemming}, but they involve knowledge or task settings from other specific domains or tasks (e.g., medical domain, chemical domain, hierarchical clustering algorithms, etc.), so they are not applicable and comparable in legal inconsistency analysis. 
To sum up, legal judgment consistency analysis over large-scale data remains to be explored.

\subsection{Legal AI}
Recently, owing to a large number of high-quality legal textual data, many researchers explore to employ NLP technology to help lawyers and other practitioners in the legal field. There are many works on generating the court's view to interpret charge results~\citep{ye2018interpretable}, retrieving relevant cases and law articles~\citep{chen2013text,raghav2016analyzing}, legal information extraction~\cite{shen2020hierarchical,chen2020joint}, legal question answering~\citep{zhong2020jec} and legal debate dialogue summarization~\citep{duan2019legal}. 

Besides, many efforts have been devoted to predicting judgment according to fact description~\citep{zhong2018legal,hu2018few,he2019secaps,chen2019charge,zhongiteratively}. These works show that neural models can effectively learn the judgment patterns from large-scale data and thus reflect the data bias, which provides the technical basis for the quantitative study. However, few researchers explore to evaluate the judgment consistency in large-scale realistic data with Legal AI technology. To the best of our knowledge, we are the first to analyse the judgment consistency in real-world data with LJP models.

\section{Problem Formulation and Methodology}
In this section, we will introduce the problem formulation and the definition of the proposed metric \metricname{}.

\subsection{Problem Formulation} 

Here we describe the problem formulation of our task in this paper.
In this paper, our task is to calculate the degree of disagreement of court decisions between $K$ different groups.
Let $a$ denotes the independent variable, which is a discrete integer ranging from $1$ to $K$, describing the feature (e.g., region, gender) used to group the dataset. Let $y$ denotes the term of penalty, and $x$ denotes the fact description. Each case $c$ can be represented as a triplet $(a^{(c)}, x^{(c)}, y^{(c)})$. We use $f(\cdot)$ to denote the judgment function, which takes the fact description $x$ as inputs and outputs the term of penalty $y$.

Given a set of legal cases $C$, we can divide it into $K$ groups $\{C_1, ..., C_K\}$ according to the value of $a$, where for each case $c \in C_i$, $a^{(c)} = i$. Then, the task is to calculate the degree of judicial disagreement between these $K$ groups.

It is notable that we can treat any feature as independent variables $a$ and thus analyze fairness from multiple perspectives. For instance, in the following experiments, we let $a$ represents the source region of cases or the gender of defendants.

\subsection{Legal Inconsistency Coefficient}

\begin{figure}[ht]
    \centering
    \includegraphics[width=1.0\columnwidth]{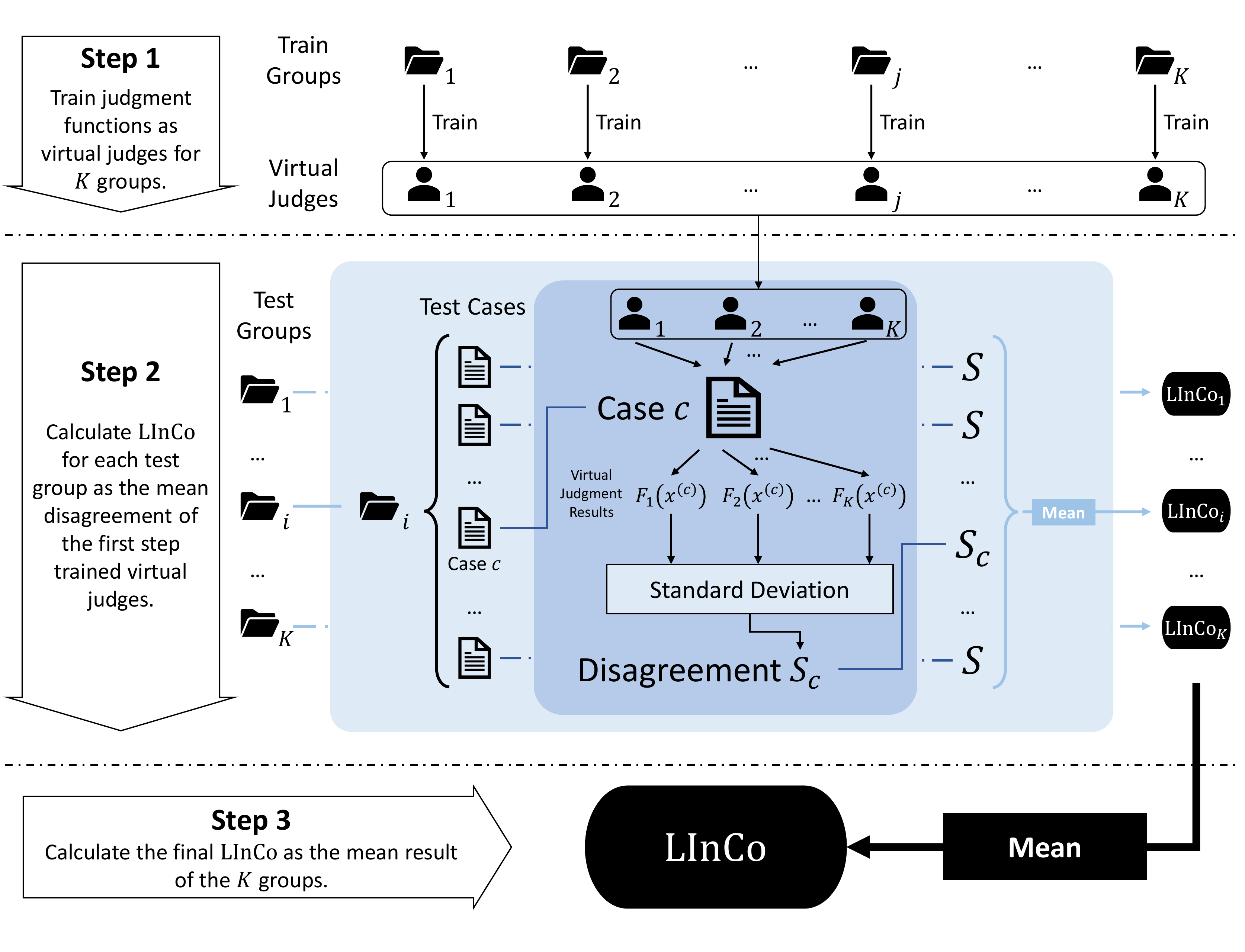}
    \caption{The overview of \metricname{}. We divide the dataset into $K$ groups, train $K$ virtual judges, and measure their disagreement to evaluate the inconsistency.}
    \label{fig:metric}
\end{figure}


In this section, we define the metric \metricname{}, the \textbf{L}egal \textbf{In}consistency \textbf{Co}efficient for measuring judicial inconsistency. Figure~\ref{fig:metric} illustrates the computation. We first train neural models to estimate the judgment function of different groups.
Then we calculate \metricname{} as the average of the disagreement on the cases in test sets.

\textbf{Estimation of Judgment Functions.} We first train LJP models to estimate the judgment functions for different groups. The judgment function $f_i(\cdot)$ can be seen as a virtual judge, which obeys both the fair judgment rules and judgment bias for group $i$. With these functions, we can estimate the results for all cases following rules and biases in different groups. We train our models to predict the term of penalty and treat this task as a regression problem. Given a case whose fact description consists of $l$ words, we first adopt an encoder (BERT~\citep{devlin2019bert} or any other models) to encode the fact description into a hidden vector. And then, a linear layer is employed to calculate the term of penalty $\hat{y}$. We use log-scaled mean square error as the loss function for optimization:
\begin{equation}
\label{equ:loss_p}
    \mathcal{L}=\left(\log y-\log \hat{y}\right)^2.
\end{equation}
We split each group $C_i$ into a training set and a test set. The training sets are used to train models, and the test sets are used to calculate \metricname{}. We can get $K$ different models with $K$ different training sets. To minimize impact from other factors that may affect consistency, we balance all train sets to have the same number of cases.

\textbf{Definition of \metricname{}.} After obtaining the judgment functions for different groups, we utilize these models to calculate \metricname{}. We define the \metricname{} as the average disagreement of all cases in the test sets: 

\begin{equation}
    \text{\metricname{}} = \frac{1}{\sum_{i=1}^{K} \lvert C^{\text{test}}_i\rvert}\sum_{i=1}^{K}\sum_{c \in C^{\text{test}}_i}S_c ,
\end{equation}
where $S_c$ is the disagreement on the case $c$. 
We argue that if the legal judgment is consistent across different groups, the judgment functions should output close, or even identical, results for the same case, and vice versa. In other words, if the judgment is consistent, the disagreement $S_c$ for the case $c$ is supposed to be close to $0$.
Therefore, we formally define $S_c$ as the standard deviation of the results of $K$ virtual judges:
\begin{equation}
    S_{c}=\sqrt{\frac{1}{K}\sum^K_{i=1}\Bigg(F_i(x^{(c)})-\frac{1}{K}\sum^K_{j=1} F_j(x^{(c)})\Bigg)^2}, 
\end{equation}
with which we can compute the disagreement for all cases. Here $F\left(\cdot\right)$ represents the standardized prediction result, which to make $S_c$ comparable across different test groups. Let $(\mu,\sigma^2)$ be the mean and variance of the ground truth $y$ in the test group, and then the standardization can be formalized as:
\begin{equation}
\begin{aligned}
    &F(x^{\left(c\right)})=\frac{f\left(x^{\left(c\right)}\right)-\mu}{\sigma}.
\end{aligned}
\end{equation}


Intuitively, models trained on a particular group can represent the characteristics and biases of that group and make judgments close to the reality of that group. By calculating the average disagreement of these ``virtual judges'' on the test set, we can measure the inconsistency of the $K$ groups. 

\section{Experiments}
In this section, we first verify the effectiveness of \metricname{} through a simulation study (Section~\ref{section:experiment1}). Then we conduct experiments on realistic datasets to analyze the judgment inconsistency in the Chinese legal system (Section~\ref{section:experiment2}). Moreover, we employ \metricname{} to evaluate the performance of several de-biasing strategies (Section~\ref{section:experiment3}).

\subsection{Dataset}
In all experiments, we construct the datasets, including synthetic dataset and real-world dataset, based on CAIL~\citep{xiao2018cail2018}, which is the largest dataset for legal judgment prediction (LJP). This dataset consists of $2.6$ million legal cases, including fact description, applicable law articles, charges, and the term of penalty published by the Supreme People’s Court of China\footnote{\url{https://wenshu.court.gov.cn/}}. 
CAIL only includes criminal cases. According to the Chinese Criminal Law, criminal judgments should be uniform across the country, so CAIL cases are supposed to be consistent theoretically. 
Based on this dataset, we construct synthetic data and real-world data for evaluation. Statistics of the dataset for every single experiment can be found in the Appendix.

Moreover, we have several settings applied to all experiments in this paper:
(1) Datasets for each group are balanced to the same size.
(2) For each dataset, we randomly select $20\%$ of data for testing and leave the remainder for training.

Please refer to the following sections for the details of synthetic and real-world datasets.

\subsection{Simulation Study}
\label{section:experiment1}

To verify the effectiveness of \metricname{}, we propose to construct synthetic datasets with controlled perturbation on the cases' term of penalty. Notably, it is a good idea to employ a crowd-sourced human evaluation to help our verification. However, the main reason for the judgment inconsistency is that the judges cannot be completely consistent. 
If we introduce a crowd-sourced human evaluation, we cannot make sure that different annotators judge consistently across different cases because they are just professionals like the real judges. 

Therefore, we do a simulation study instead, in which the inconsistency of different groups is well controlled. Thus, we can evaluate the reliability of \metricname{} by computing the correlation between \metricname{} and the controlled inconsistency.

\subsubsection{Experimental Settings}
\textbf{Dataset Construction.}
We construct the synthetic data of different groups by keeping facts the same and perturbing the term of penalty. In other words, each data group uses the same set of facts, but the ground truths are perturbed for each group. To simulate the inconsistency between groups, we make the perturbations vary for each data group.

Let $\beta$ denote the inconsistency factor of the synthetic dataset. We build a synthetic dataset with the following steps:

(1) We first randomly select $N$ cases $O = \{(x^{\text{o}}_1, y^{\text{o}}_1), \dots, (x^{\text{o}}_N, y^{\text{o}}_N)\}$ from the CAIL dataset.

(2) Then, for the $i$-th case $(x_i^g, y_i^g)$ in the group $C_g$, we keep the facts the same as those in $O$ and perturb the term of the penalty:
\begin{equation}
    x_i^g = x^{\text{o}}_i, \quad y_i^g = \max \{ 0, \gamma_{g,i} y^{\text{o}}_i\}.
\end{equation}
Here $\gamma_{g,i}$ is randomly sampled following the normal distribution $N(\mu_g, \sigma_g^2)$, where
\begin{equation}
    \mu_g = 1-\beta +\frac{2g\beta}{K-1}, \quad \sigma_g = \frac{\lvert 1-\mu_g \rvert}{3}.
\end{equation}

Intuitively, it can be considered that we give each group a scaling $\mu_i$, and apply an additional random effect to each case. The values of $\mu_g$ make the single-group scaling roughly in the range $[1-\beta, 1+\beta]$, and the value of $\sigma_g$ makes $\gamma$ fall between $1$ and $2\mu-1$ with a $99.7\%$ probability.

We argue that if \metricname{} is sufficient to evaluate the legal judgment inconsistency, the correlation coefficient between \metricname{} and the inconsistency factor $\beta$ is supposed to be close to $1$.

\noindent
\textbf{Models.}
We employ two strategies to estimate the judgment functions for different groups.
\begin{itemize}[leftmargin=*]
	\item \textbf{DGN.} We adopt DGN~\citep{chen2019charge}, which is designed for predicting the term of penalty, as our encoder to train virtual judges. Compared with general NLP models, DGN mainly focuses on the relationship between charge and term. DGN stacks multiple blocks of an LSTM layer and a charge-specific gating layer for generating a focused charge-based representation of the case. Therefore, this work can make good use of charge information to help judgment prediction.
	\item \textbf{Golden Value.} As mentioned before, \metricname{} represents the disagreement of models to reflect the inconsistency. However, \metricname{} will be influenced by both the judgment inconsistency and model misspecification.
	Therefore, we propose to use the synthetic ground truth to calculate the golden value of \metricname{} for comparison. More specifically, when given the case $c$, the virtual judge $f_g$ will output the ground truth value $y_g^{\left(c\right)}$ in group $g$.
\end{itemize}

\noindent
\textbf{Training Settings.}

We use trainable character-level embeddings for DGN and use Adam~\citep{kingma2014adam} to train all models. 
The learning rate is $10^{-3}$. We implement DGN using PyTorch. You can find the codes in the attached files.




\subsubsection{Experimental Results and Analysis}

\begin{figure}[ht]
    \centering
  \includegraphics[width=0.95\columnwidth]{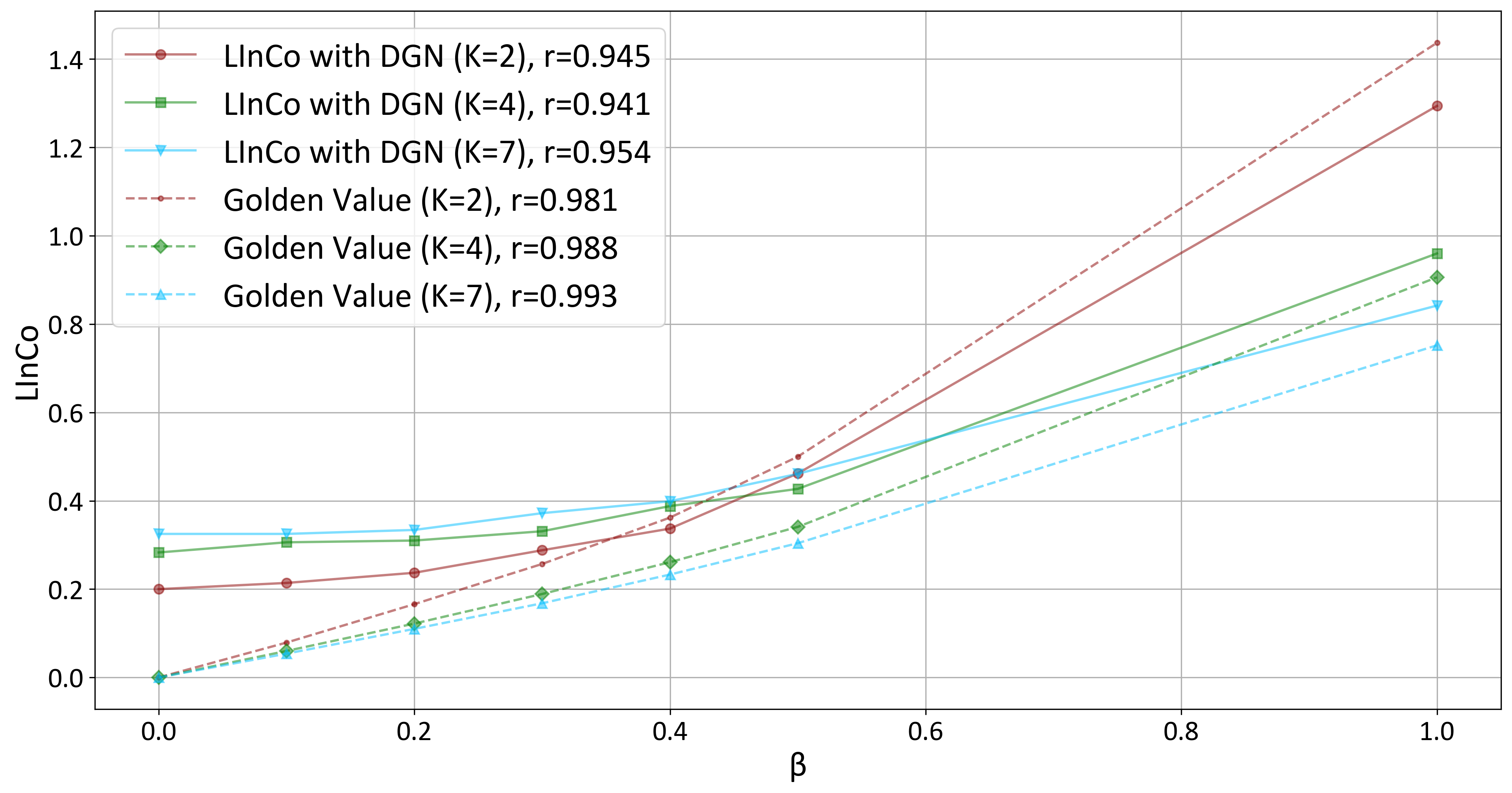}
    \caption{The curves in the figure show the tendency of \metricname{} to change with the increasing of the inconsistency factor $\beta$. Here $r$ represents the correlation coefficient between $\beta$ and $\metricsymbol$. For each result, we run $7$ times for average (this setting has been applied to all experiments in this paper).}
    \label{fig:exp1_table_line}
\end{figure}


The main results of the simulation study are shown in Figure~\ref{fig:exp1_table_line}. Please refer to the Appendix for the detailed values for each point in the figure. From the results, we can observe that: 

(1) When $K$ is fixed, a larger inconsistency factor $\beta$ will lead to larger \metricname{} as expected. We also calculate the correlation coefficient between $\beta$ and \metricname{}. The correlation coefficients for the golden value and \metricname{} and DGN are relatively high, which proves that \metricname{} is reliable in theory and in practice.
Thus \metricname{} can be used to evaluate the judicial inconsistency in the real world.

(2) There is a difference between the \metricname{} with DGN and the golden value. For example, When $\beta=0$, which means the judgment is consistent across all groups, the \metricname{} with DGN is greater than $0$. This indicates that neural models will introduce bias due to the model misspecification. In reality, the model-introduced bias is inevitable because we are unable to know the underlying perfect model for prediction. However, the difference between golden value and \metricname{} with DGN is small, which means the model misspecification has little impact on \metricname{}. And the correlation coefficients between $\beta$ and \metricname{} with DGN are greater than $0.94$ with an average of $0.955$, which further proves the robustness of \metricname{}.

\subsection{\metricname{} on Real-World Dataset}
\label{section:experiment2}

In this section, we analyze the judgment inconsistency on CAIL with \metricname{}. We mainly focus on regional inconsistency and gender inconsistency. We match all data with cases on China Judgment Online\footnote{\url{ http://wenshu.court.gov.cn/}} to obtain the source region and the defendant's gender for each case. On one hand, when studying the regional consistency, we select data from $K=7$ different provincial-level administrative regions\footnote{The provincial-level administrative region is the first-level administrative division unit of China. In China, there are 34 provincial-level administrative regions, including 23 provinces, 5 autonomous regions, 4 municipalities, and 2 special administrative regions.} and use the data from these regions as the $7$ groups for the experiments. For anonymity reasons, we do not list the names of these regions. On the other hand, for the gender inconsistency, we set $K=2$ and group the data by the defendant's gender assignment.

We first conduct experiments on nine crimes to explore the relationship between legal judgment consistency and the severity of crimes. Detailed descriptions of these charges can be found in the Appendix. In the Chinese legal system, the severity of individual cases is divided into three levels: sentences being imprisonment of fewer than $3$ years, $3$ to $10$ years, and more than $10$ years. Therefore, we calculate the proportion of each crime's actual judgment falling in these three ranges to reflect the severity of the charge.

\begin{table}[!t]
\small
    \centering
    \caption{Results of \metricname{} on nine crimes, compared with the distributions of real judgment for these crimes. The distributions reflect the crimes' severity.} 
    \label{table:crime_result}
    \tabcolsep 20pt
    \begin{tabular*}{\textwidth}{ccccc}\toprule
\multicolumn{3}{c}{\textbf{Real-world Sentence Distribution}} & \multirow{2}{*}{\textbf{Crimes}} & \multirow{2}{*}{\textbf{\metricname{}}} \\ \cline{1-3}
        $\mathbf{\left[0,3\right)}$ & $\mathbf{\left[3,10\right)}$ & $\mathbf{\left[10,\infty\right)}$ &  &  \\ \hline
        55\% & 34\% & 11\% & Fraud & 0.457 \\ 
        67\% & 25\% & 8\% & Drug Trafficking & 0.487 \\ 
        73\% & 22\% & 5\% & Possession of Illegal Drugs & 0.541 \\ 
        84\% & 14\% & 2\% & Intentional Injury & 0.274 \\ 
        86\% & 14\% & 0\% & Traffic Offence & 0.553 \\ 
        95\% & 5\% & 0\% & Theft & 0.376 \\ 
        97\% & 3\% & 0\% & Picking Quarrels and Provoking Trouble & 0.698 \\ 
        100\% & 0\% & 0\% & Disrupting Public Service & 0.725 \\ 
        100\% & 0\% & 0\% & Providing Venues for Drug Users & 0.870 \\
        \bottomrule
    \end{tabular*}
\end{table}

The results are shown in Table~\ref{table:crime_result}. We notice that the judicial inconsistency is negatively correlated with the severity of the selected crime in general. In other words, overall, the sentences for the more serious crimes are more consistent, which indicates that the judges may be more careful about felony trials than misdemeanor ones.

\begin{table}[ht]
        \centering
        \small
        \tabcolsep 36pt
        \caption{\metricname{} calculates on crimes with sufficient data. All data is from one particular region and divided into $K=7$ groups. Results for cross-region inconsistency is also shown for comparison.
    }
    \label{table:same_region}
        \begin{tabular*}{\textwidth}{ccc}\toprule
             \textbf{Crime} & \textbf{Cross-region} & \textbf{Single region} \\\hline
    Fraud & 0.457 & 0.371 \\ 
    Drug Trafficking & 0.487 & 0.304 \\ 
    Traffic Offence & 0.553 & 0.127 \\ 
    Theft & 0.376 & 0.262 \\ 
    \tabincell{c}{Picking Quarrels and Provoking Trouble} & 0.698 & 0.145 \\
    \bottomrule
        \end{tabular*}
    
\end{table}

To demonstrate that region is indeed the main factor influencing the results in Table~\ref{table:crime_result}, we calculate the inconsistency without regional differences for comparison. We select several crimes with sufficient data and randomly divide the data from one particular region into $K=7$ groups for the experiments. The results of \metricname{} are shown in Table~\ref{table:same_region}. It can be observed that \metricname{} within one single region is much smaller than \metricname{} between different regions. This reveals that region is an important factor for the inconsistency in Table~\ref{table:crime_result}. Besides, some other factors may have an impact on the consistency like GDP and education levels, especially for some of the crimes like fraud or theft.



\begin{table}[ht]
    \centering
    \caption{Regional legal judgment inconsistency between different years.}
    \label{table:diff_year_regional}
    \small
        \tabcolsep 36pt
        \begin{tabular*}{\textwidth}{ccc}\toprule
             \textbf{Crime} & \textbf{2013 -- 2015} & \textbf{2016 -- 2018} \\\hline  
    Fraud & 0.504 & 0.516 \\ 
    Drug Trafficking & 0.523 & 0.518 \\ 
    Intentional Injury & 0.305 & 0.285 \\ 
    Traffic Offence & 0.627 & 0.629 \\ 
    Theft & 0.356 & 0.340 \\ 
    \tabincell{c}{Picking Quarrels and Provoking Trouble} & 0.738 & 0.755 \\
    \bottomrule
        \end{tabular*}
\end{table}

Moreover, we conduct experiments on data in different years in an attempt to verify whether regional judgment inconsistency changes over time, and the results are shown in Table~\ref{table:diff_year_regional}. It can be observed that the regional inconsistency is nearly the same across the years, which indicates that inconsistency does not change significantly over time.

\begin{table}[ht]
    \centering
    \small
    \tabcolsep 20pt
    \caption{Results of gender inconsistency, compared with the distributions of real judgment for these crimes.}
    \label{table:gender_result}
\begin{tabular}{ccccc}\toprule
\multicolumn{3}{c}{\textbf{Real-world Sentence Distribution}} & \multirow{2}{*}{\textbf{Crimes}} & \multirow{2}{*}{\textbf{\metricname{}}} \\ \cline{1-3}
        $\mathbf{\left[0,3\right)}$ & $\mathbf{\left[3,10\right)}$ & $\mathbf{\left[10,\infty\right)}$ &  &  \\ \hline
        55\% & 34\% & 11\% & Fraud & 0.216 \\ 
        67\% & 25\% & 8\% & Drug Trafficking & 0.182 \\ 
        73\% & 22\% & 5\% & Possession of Illegal Drugs & 0.258 \\ 
        84\% & 14\% & 2\% & Intentional Injury & 0.163 \\ 
        86\% & 14\% & 0\% & Traffic Offence & 0.251 \\ 
        95\% & 5\% & 0\% & Theft & 0.220 \\ 
        97\% & 3\% & 0\% & Picking Quarrels and Provoking Trouble & 0.351 \\ 
        100\% & 0\% & 0\% & Disrupting Public Service & 0.353 \\ 
        100\% & 0\% & 0\% & Providing Venues for Drug Users & 0.338 \\
        \bottomrule
    \end{tabular}
\end{table}

We also calculate \metricname{} to detect gender inconsistency, and the results are shown in Table~\ref{table:gender_result}. Notably, sexual harassment or other gender-sensitive charges are not included, so that there should be complete equality between men and women. From the table, we can observe that the inconsistency is negatively correlated with the crime's severity generally persists. Furthermore, we can find that gender inconsistency is much less than regional inconsistency, which means biased sentencing due to gender discrimination is rare, or at least less than that due to regional differences.

    




\subsection{De-biasing Strategies Evaluation}
\label{section:experiment3}

In the previous section, we have defined \metricname{} for measuring judicial inconsistency with the help of LJP, and we find that inconsistency does exist in real datasets. 

However, inconsistency should be avoided in both the real-world judgment and for the LJP model. Therefore, in this section, we use \metricname{} to evaluate the de-bias performance of several de-biasing methods. We will introduce two training strategies for de-biasing and compare them with the regular one. As in the previous section, we focus on regional inconsistency as an example.


\subsubsection{Training Strategies and Optimization}\label{section:training}

\begin{figure*}[t]
    \centering
    \includegraphics[width=0.95\textwidth]{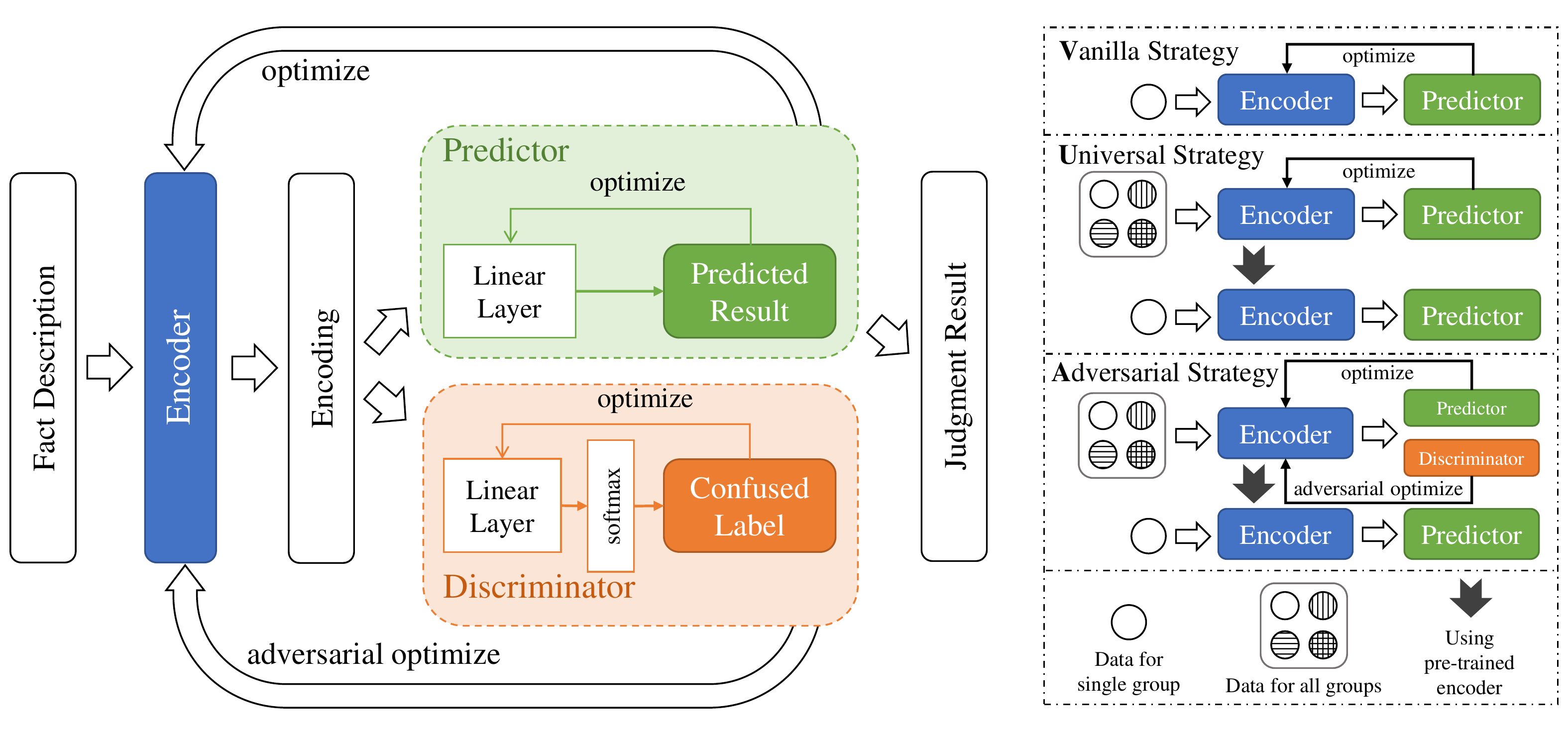}
    \caption{An overview of the training framework and three strategies.}
    \label{fig:framework}
\end{figure*}

In this part, we describe the three different training strategies. To summarize: Vanilla Strategy fully follows the method of LJP task to train $K$ virtual judges with regional differences; Universal Strategy tries to reduce regional differences by pre-training a shared encoder over regions;  Adversarial Strategy is similar to Universal Strategy except that an adversarial pre-trained encoder is used to eliminate regional differences further. An overview of all strategies can be found in Figure~\ref{fig:framework}.


\textbf{Vanilla Strategy}.
Vanilla Strategy (V-Strategy) only utilizes encoder and predictor, which means it is equal to the traditional LJP model and does not do any de-biasing at all.



\textbf{Universal Strategy}.
The same as V-Strategy, Universal Strategy (U-Strategy) utilizes encoder and predictor. The difference is that we first use the entire dataset to pre-train an encoder and then train the $K$ models separately with the fixed pre-trained encoder (only optimize the predictor) on each train set. By this strategy, models can share the parameters of embedding for de-biasing.




\textbf{Adversarial Strategy}. 
Adversarial Strategy (A-Strategy) is similar to U-Strategy, except that we introduce a discriminator when pre-training the encoder. 
The discriminator is used for predicting which group the input data belongs to, i.e., from which region, and we can formalize the task of discriminator as a multi-class classification problem. We use a linear layer in the discriminator and use cross-entropy to optimize the models in our experiments.
Specifically, we use $\mathcal{L}_\text{D}$ to optimize the discriminator's linear layer and use $\mathcal{L}_\text{A}$ to adversarial optimize the encoder as:
\begin{equation}
\begin{aligned}
    \mathcal{L}_\text{D}=-\sum\limits_{i=1}^{K} \Big(\left[ a=i\right]\log {\pi_{i}} + \big(1-\left[ a=i\right]\big)\log {\big(1-\pi_{i}\big)} \Big), \\
    \mathcal{L}_\text{A}=-\sum\limits_{i=1}^{K} \Big(\big(1-\left[ a=i\right]\big)\log {\pi_{i}} + \left[ a=i\right]\log {\big(1-\pi_{i}\big)} \Big).
\end{aligned}
\end{equation}
Here $\pi_i \in \left[ 0,1 \right] $ ($1\leq i\leq K$) indicates the probability of $a=i$ output by the model.
For training, we first use the entire dataset to train the model with a discriminator, which targets to obtain the pre-trained encoder that can confuse the discriminator.
Then we train $K$ independent virtual judges with this pre-trained encoder, which is the same as U-Strategy.

\subsubsection{Experimental Settings}

In the experiments of verifying the strategies, we mainly follow the experimental settings mentioned in previous experiments.
In addition, besides the specific term predicting model DGN, we employ more NLP models for classification and regression here. 

\textbf{DPCNN}~\citep{johnson2017deep}: This method is based on Convolutional Neural Networks. It uses region embedding layers and convolutional blocks, combined with shortcut connections proposed in ResNet~\citep{he2016deep}. The structure of DPCNN makes deep models for NLP classification tasks possible, and it can effectively extract distant relationship features in the text. 

\textbf{GRU}~\citep{cho2014learning}: This model is a variation of LSTM. GRU adds a gating mechanism in the recurrent neural network unit, which couples the input and forget gates to decrease the number of parameters. Even with fewer parameters, GRU could still achieve similar performances as LSTM.

\textbf{BERT}~\citep{devlin2019bert}: BERT is the model formed by multiple bidirectional Transformer layers. The parameters of BERT has been fully pre-trained on large-scale text corpora. 
Recently, BERT has achieved state-of-the-art in many NLP tasks, including classification, reading comprehension, and question answering.

To ensure a fair comparison between different models, we use trainable character-level embeddings for every model.  
We use Adam~\citep{kingma2014adam} to train all models except BERT, for which we use BertAdam~\citep{devlin2019bert}. 
The learning rate is $10^{-5}$ for BERT, $10^{-2}$ for DPCNN, and $10^{-3}$ for all other models. 
As above, we implement all the models using PyTorch, and you can find the codes in the attached files.

\subsubsection{Inconsistency Evaluation with Different Strategies}

In this part, we present the experimental results of the three strategies. We evaluate \metricname{} of these strategies using different models on the real dataset. We pick the two charges with the most data, i.e., theft and intentional injury, to prevent models overfitting on small datasets.

\begin{table}[ht]
    \centering
    \small
    \tabcolsep 21.5pt
    \caption{Results of \metricname{} on two representative crimes using three strategies.}
\label{table:result_above}
    \begin{tabular*}{\textwidth}{c|ccc|ccc}\toprule
         \textbf{Crimes} & \multicolumn{3}{c|}{\textbf{Theft}}  & \multicolumn{3}{c}{\textbf{Intentional Injury}}  \\\hline
         \textbf{Strategy} & \textbf{V} & \textbf{U} & \textbf{A} & \textbf{V} & \textbf{U} & \textbf{A} \\ \hline
DGN &
0.376 & 0.056 & 0.128 & 0.274 & 0.053 & 0.113


\\
DPCNN &
0.307 & 0.191 & 0.175 & 0.255 & 0.127 & 0.119
\\
GRU &
0.298 & 0.084 & 0.117 & 0.219 & 0.076 & 0.084
\\
BERT &
0.305 & 0.194 & 0.175 & 0.251 & 0.227 & 0.16
\\
\bottomrule
    \end{tabular*}
\end{table}

We calculate \metricname{} for all models in each strategy, shown in Table~\ref{table:result_above}. From the table, we find that overall, models trained with the U-Strategy and A-Strategy have lower \metricname{} than those with V-Strategy.
It is also worth noting that the de-biasing effect of U-Strategy is better than A-Strategy for the two sequential models, DGN and GRU. In contrast, A-Strategy is better for the two non-sequential models. This phenomenon suggests that we need to use different strategies to de-bias different models.






\begin{figure*}[t]
    \centering
  \includegraphics[width=0.95\textwidth]{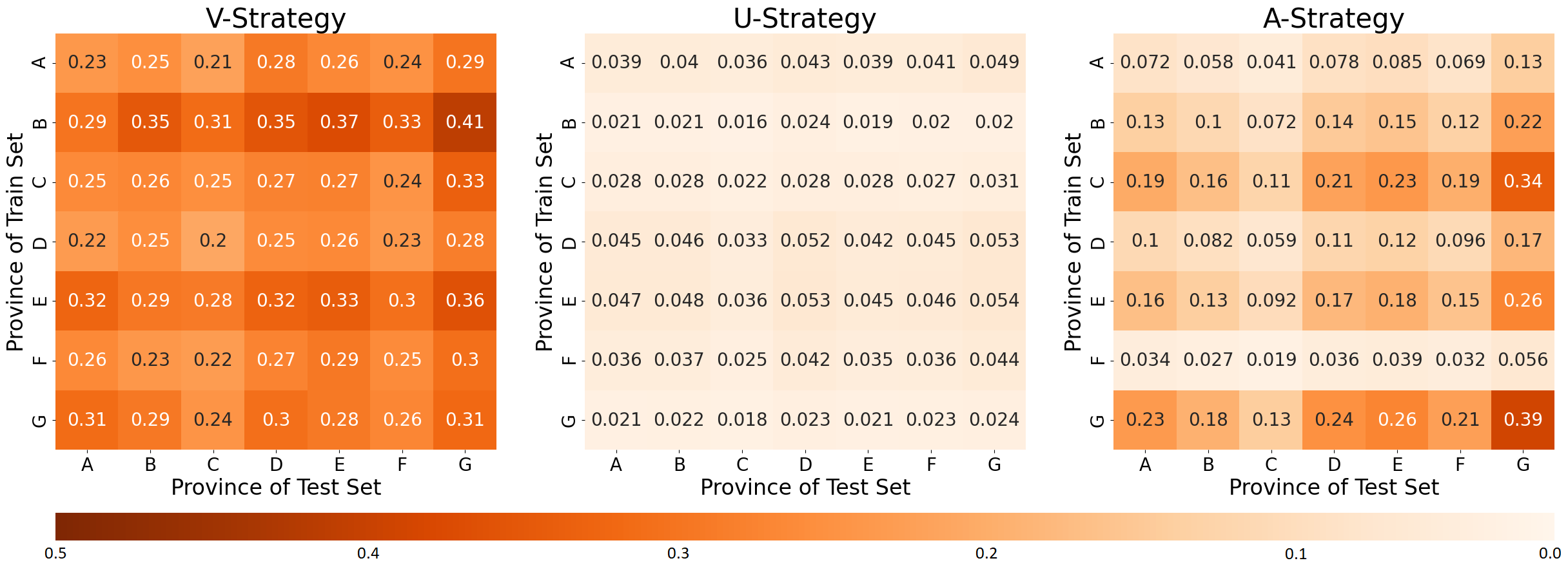}
    \caption{Heat maps of $M$ with DGN as encoder on the crime of theft.}
    \label{fig:heatmap}
    \vspace{-1em}
\end{figure*}

To better interpret the results, we propose and compute matrix $M\in\mathbb{R}^{K\times K}$ to measure inconsistency among $K$ virtual judges for visualization, as the following equation:
\begin{equation}
\begin{aligned}
    M_{i,t}=\frac{1}{\lvert  C^{\text{test}}_i\rvert}\sum_{c \in C^{\text{test}}_i}\Bigg\lvert F_i(x^{(c)})-\frac{1}{K}\sum^K_{j=1} F_j(x^{(c)})\Bigg\rvert .
\end{aligned}
\end{equation}
Here $t$ represents the index of the test set. This matrix can intuitively and roughly reflect the trend of \metricname{}. We choose DGN as the encoder and the crime of theft and visualize the heat map of $M$ in Figure~\ref{fig:heatmap}.
From the figure, we can see that the inconsistency of the three strategies when using DGN corresponds to our analysis of the sequential model, i.e., V-Strategy has the worst consistency, and U-Strategy has the best consistency.

From the figure we can find that for three different strategies, the inconsistency of the model predictions gradually decreases as we expected, which reflects: (1) Encoders inevitably catch regional diversity during the training process of V-Strategy; (2) U-Strategy tends to extract general features instead of specific regional features, which improves prediction consistency; (3) A-Strategy almost eliminates the regional diversity learned by the entire network, thus even when train sets from different regions are used, the network can still maintain high consistency in prediction.
In addition, we can further confirm \metricname{} can directly reveal judgment inconsistency quantitatively and accurately.

\begin{table}[ht]
    \centering
    \small
    \tabcolsep 23pt
    \caption{The average L1 distance between ground truth and prediction of models on two crimes (unit: month).}
    \label{table:models_deviation}
    \begin{tabular*}{\textwidth}{c|ccc|ccc}\toprule
         \textbf{Crimes} & \multicolumn{3}{c|}{\textbf{Theft}}  & \multicolumn{3}{c}{\textbf{Intentional Injury}}  \\\hline
         \textbf{Strategy} & \textbf{V} & \textbf{U} & \textbf{A} & \textbf{V} & \textbf{U} & \textbf{A} \\ \hline
DGN &
6.74 & 5.18 & 7.40 & 8.57 & 7.60 & 9.39
\\
DPCNN &
6.22 & 7.95 & 8.30 & 9.44 & 11.35 & 13.72
\\
GRU &
6.27 & 5.05 & 6.80 & 7.58 & 6.90 & 9.58
\\
BERT &
5.48 & 5.01 & 7.76 & 8.08 & 7.53 & 12.53
\\
\bottomrule
    \end{tabular*}
\end{table}

We also pay attention to the performance of each network for each strategy. Specifically, we calculate the average L1 distance between prediction and ground truth to measure the accuracy performance of each strategy. The results can be found in Table~\ref{table:models_deviation}. It indicates that U-Strategy often leads to the best performance and A-Strategy to the worst. U-Strategy is usually the best because it learns from more data, so the encoders are strong enough to remember the features of every region to reach a better performance. However, since A-strategy will eliminate these regional features, it understandably results in worse performance as the judgment results between regions are inherently inconsistent.

As a result, if we want to apply LJP technology to the real legal system, we must use some strategies to reduce the inconsistency of models. Otherwise, we cannot promise the fairness of the legal system. On the other hand, the model cannot perform too poorly to be used in the system. So our strategy should achieve a balance between inconsistency and performance (like U-strategy) to be applied to the real world.

\section{Discussion}\label{section:discussion}

Our experimental results show that inconsistency does exist in real datasets. However, many possible factors may lead to inconsistency, such as regional differences in economic conditions, education levels, differences between courts or judges, interference of public opinions, etc. The reasons for inconsistency may be very complicated, and we will leave it as our future work.




Besides, we list several questions and answers in the Appendix for readers to understand our paper better. Here we show the six most typical questions as following:

\textbf{Q: Measuring judgment inconsistency is already difficult for humans. Why to automatically do it?}
A: Humans cannot accurately perform macro-analysis due to the huge amount of data, which is the specialty of computers.

\textbf{Q: What is the motivation of \metricname{}?}
A: The motivation of \metricname{} can be illustrated with the following example. If we want to measure regional consistency, the best way is to have local judges independently adjudicate the same set of cases and observe their disagreement. However, this is unrealistic, so we train LJP models to simulate local virtual judges to achieve similar results.

\textbf{Q: Since judgment consistency means the degree to which similar cases yields similar court decision, why not measure the similarity between cases to help solve the problem?}
A: There are two reasons. (1) Facing $K$ groups with $N$ cases in each group, the time complexity of measuring the similarity between cases will be $\mathrm{O}\left(\left( K N \right)^2\right)$, which is really time-consuming. (2) It is hard to define the similarity between two different cases~\citep{xiao2019cail2019-scm:,zhong2020does}. For example, an act of self-defense can make two textually similar cases very different, and much more factors should be considered when evaluating the similarity between cases.

\textbf{Q: For the LJP model, why do we select the term of penalty rather than charges or applicable articles as the task?} 
A: 
Both relevant charges, applicable articles, and the term of penalty are the task of LJP. However, the first two are qualitative, while the term is quantitative. Comparatively, qualitative judgment results are less likely to be inconsistent. For example, it is rare for a case that should have been convicted of theft to be convicted of fraud. In contrast, inconsistencies are often reflected in the term of penalty. Even similar cases may have various lengths of penalty, as shown in the first table in the main text. Therefore, the term of penalty can better reflect consistency, and it is why we select it as the major task.

\textbf{Q: Why not do a case study?}
A:
\metricname{}, as a macro indicator, presents the overall inconsistency of a certain amount of data. In contrast, case studies show differences between individual cases, so it makes no sense in this paper.

\textbf{Q: Why not use unsupervised methods, such as clustering, to solve the problem?}
A:
Traditional clustering methods can measure consistency between texts. However, our study focuses more on judgment, which is the intermediate process from the factual text to the judgment result. It is hard to represent the judgment process in terms of vectors, so clustering cannot address the issue. Besides, clustering cannot learn the document representation well and is therefore too weak to be applied.



\section{Conclusions}

In this paper, we address the lack of methods for measuring the inconsistency of legal judgment and propose \metricname{}, an approach that quantifies the judgment inconsistency between different data groups.
We verify the effectiveness of \metricname{} on the synthetic datasets and make sure that \metricname{} can be applied to measure legal judgment inconsistency.
Then, we conduct experiments on the real-world datasets and discover that judicial inconsistency indeed exists in the Chinese legal system.
We also discover several interesting phenomena, including that felonies tend to be judged more consistently than misdemeanors, and gender inconsistency is much less than the regional one.
Moreover, we use \metricname{} to evaluate the performance of several de-biasing strategies (such as adversarial learning).

We will focus on exploring the following directions in the future:
(1) We will explore the relationship between judicial inconsistency and other factors and discover patterns in the legal system with the help of \metricname{}. 
(2) We will also investigate which factors cause legal inconsistency, which will be important for improving the fairness of the judicial system.

We hope that with more research on judgment consistency analysis, judicial fairness and consistency can be better quantified so that the entire legal system can be better monitored, and the development of judgment fairness can be promoted.

\Acknowledgements{This work was supported by National Natural Science Foundation of China (Grant Nos. 00000000 and 11111111).}





\bibliographystyle{plainnat}
\bibliography{anthology}

\begin{thebibliography}{34}
\providecommand{\natexlab}[1]{#1}
\providecommand{\url}[1]{\texttt{#1}}
\expandafter\ifx\csname urlstyle\endcsname\relax
  \providecommand{\doi}[1]{doi: #1}\else
  \providecommand{\doi}{doi: \begingroup \urlstyle{rm}\Url}\fi

\bibitem[Anderlini et~al.(2014)Anderlini, Felli, and Riboni]{anderlini2014why}
Luca Anderlini, Leonardo Felli, and Alessandro Riboni.
\newblock Why stare decisis.
\newblock \emph{Review of Economic Dynamics}, 17\penalty0 (4):\penalty0
  726--738, 2014.

\bibitem[Anderlini et~al.(2020)Anderlini, Felli, and
  Riboni]{anderlini2020legal}
Luca Anderlini, Leonardo Felli, and Alessandro Riboni.
\newblock Legal efficiency and consistency.
\newblock \emph{European Economic Review}, 121:\penalty0 103323, 2020.

\bibitem[Berk et~al.(2018)Berk, Heidari, Jabbari, Kearns, and
  Roth]{berk2018fairness}
Richard Berk, Hoda Heidari, Shahin Jabbari, Michael Kearns, and Aaron Roth.
\newblock Fairness in criminal justice risk assessments: The state of the art.
\newblock \emph{Sociological Methods \& Research}, page 0049124118782533, 2018.

\bibitem[Chen et~al.(2019)Chen, Cai, Dai, Dai, and Ding]{chen2019charge}
Huajie Chen, Deng Cai, Wei Dai, Zehui Dai, and Yadong Ding.
\newblock Charge-based prison term prediction with deep gating network.
\newblock In \emph{Proceedings of EMNLP}, 2019.

\bibitem[Chen et~al.(2020)Chen, Sun, Yang, and Lin]{chen2020joint}
Yanguang Chen, Yuanyuan Sun, Zhihao Yang, and Hongfei Lin.
\newblock Joint entity and relation extraction for legal documents with legal
  feature enhancement.
\newblock In \emph{Proceedings of COLING}, pages 1561--1571, 2020.

\bibitem[Chen et~al.(2013)Chen, Liu, and Ho]{chen2013text}
Yen-Liang Chen, Yi-Hung Liu, and Wu-Liang Ho.
\newblock A text mining approach to assist the general public in the retrieval
  of legal documents.
\newblock \emph{Journal of the American Society for Information Science and
  Technology}, 64\penalty0 (2):\penalty0 280--290, 2013.

\bibitem[Cho et~al.(2014)Cho, Merrienboer, Gulcehre, Bahdanau, Bougares,
  Schwenk, and Bengio]{cho2014learning}
Kyunghyun Cho, Bart~Van Merrienboer, Caglar Gulcehre, Dzmitry Bahdanau, Fethi
  Bougares, Holger Schwenk, and Yoshua Bengio.
\newblock Learning phrase representations using rnn encoder-decoder for
  statistical machine translation.
\newblock In \emph{Proceedings of EMNLP}, 2014.

\bibitem[Devlin et~al.(2019)Devlin, Chang, Lee, and Toutanova]{devlin2019bert}
Jacob Devlin, Ming-Wei Chang, Kenton Lee, and Kristina Toutanova.
\newblock Bert: Pre-training of deep bidirectional transformers for language
  understanding.
\newblock In \emph{Proceedings of NAACL}, 2019.

\bibitem[Duan et~al.(2019)Duan, Zhang, Yuan, Zhou, Liu, Wang, Wang, Zhang, Sun,
  and Wu]{duan2019legal}
Xinyu Duan, Yating Zhang, Lin Yuan, Xin Zhou, Xiaozhong Liu, Tianyi Wang,
  Ruocheng Wang, Qiong Zhang, Changlong Sun, and Fei Wu.
\newblock Legal summarization for multi-role debate dialogue via controversy
  focus mining and multi-task learning.
\newblock In \emph{Proceedings of CIKM}, 2019.

\bibitem[Dwork et~al.(2012)Dwork, Hardt, Pitassi, Reingold, and
  Zemel]{dwork2012fairness}
Cynthia Dwork, Moritz Hardt, Toniann Pitassi, Omer Reingold, and Richard~S
  Zemel.
\newblock Fairness through awareness.
\newblock In \emph{Proceedings of ITCS}, 2012.

\bibitem[Edgely(2009)]{edgely2009common}
Michelle Edgely.
\newblock Common law sentencing of mentally impaired offenders in australian
  courts: A call for coherence and consistency.
\newblock \emph{Psychiatry, Psychology and Law}, 16\penalty0 (2):\penalty0
  240--261, 2009.

\bibitem[Flynn(2013)]{flynn2013making}
Eilion{\'o}ir Flynn.
\newblock Making human rights meaningful for people with disabilities:
  Advocacy, access to justice and equality before the law.
\newblock \emph{The International Journal of Human Rights}, 17\penalty0
  (4):\penalty0 491--510, 2013.

\bibitem[Grgichlaca et~al.(2018)Grgichlaca, Redmiles, Gummadi, and
  Weller]{grgichlaca2018human}
Nina Grgichlaca, Elissa~M Redmiles, Krishna~P Gummadi, and Adrian Weller.
\newblock Human perceptions of fairness in algorithmic decision making: A case
  study of criminal risk prediction.
\newblock In \emph{Proceedings of WWW}, 2018.

\bibitem[He et~al.(2019)He, Peng, Le, He, and Zhu]{he2019secaps}
Congqing He, Li~Peng, Yuquan Le, Jiawei He, and Xiangyu Zhu.
\newblock Secaps: a sequence enhanced capsule model for charge prediction.
\newblock In \emph{International Conference on Artificial Neural Networks},
  pages 227--239. Springer, 2019.

\bibitem[He et~al.(2016)He, Zhang, Ren, and Sun]{he2016deep}
Kaiming He, Xiangyu Zhang, Shaoqing Ren, and Jian Sun.
\newblock Deep residual learning for image recognition.
\newblock In \emph{Proceedings of CVPR}, pages 770--778, 2016.

\bibitem[Hu et~al.(2018)Hu, Li, Tu, Liu, and Sun]{hu2018few}
Zikun Hu, Xiang Li, Cunchao Tu, Zhiyuan Liu, and Maosong Sun.
\newblock Few-shot charge prediction with discriminative legal attributes.
\newblock In \emph{Proceedings of COLING}, 2018.

\bibitem[Jezewski et~al.(2003)Jezewski, Horoba, Gasek, Wrobel, Matonia, and
  Kupka]{jezewski2003analysis}
Janusz Jezewski, Krzysztof Horoba, A~Gasek, Janusz Wrobel, Adam Matonia, and
  Tomasz Kupka.
\newblock Analysis of nonstationarities in fetal heart rate signal:
  inconsistency measures of baselines using acceleration/deceleration patterns.
\newblock In \emph{Seventh International Symposium on Signal Processing and Its
  Applications, 2003. Proceedings.}, volume~2, pages 9--12. IEEE, 2003.

\bibitem[Johnson and Zhang(2017)]{johnson2017deep}
Rie Johnson and Tong Zhang.
\newblock Deep pyramid convolutional neural networks for text categorization.
\newblock In \emph{Proceedings of ACL}, 2017.

\bibitem[Kingma and Ba(2015)]{kingma2014adam}
Diederik Kingma and Jimmy Ba.
\newblock Adam: A method for stochastic optimization.
\newblock In \emph{Proceedings of ICLR}, 2015.

\bibitem[Korenius et~al.(2004)Korenius, Laurikkala, J{\"a}rvelin, and
  Juhola]{korenius2004stemming}
Tuomo Korenius, Jorma Laurikkala, Kalervo J{\"a}rvelin, and Martti Juhola.
\newblock Stemming and lemmatization in the clustering of finnish text
  documents.
\newblock In \emph{Proceedings of the thirteenth ACM international conference
  on Information and knowledge management}, pages 625--633, 2004.

\bibitem[Li(2014)]{li2014a}
Lixia Li.
\newblock A sentencing study of the criminal cases with similar conditions but
  different sentence (in chinese).
\newblock \emph{Journal of Zhongzhou University}, pages 34--37, 2014.

\bibitem[Li et~al.(2012)Li, Lin, and Chen]{li2012inconsistency}
Teng Li, Chengtao Lin, and Quanshi Chen.
\newblock Inconsistency analysis of lifepo (4) battery packing.
\newblock \emph{Journal of Tsinghua University Science and Technology},
  52\penalty0 (7):\penalty0 1001--1006, 2012.

\bibitem[Raghav et~al.(2016)Raghav, Reddy, and Reddy]{raghav2016analyzing}
K~Raghav, P~Krishna Reddy, and V~Balakista Reddy.
\newblock Analyzing the extraction of relevant legal judgments using
  paragraph-level and citation information.
\newblock \emph{Artificial Intelligence for Justice}, 30, 2016.

\bibitem[Reamer(2005)]{reamer2005ethical}
Frederic~G Reamer.
\newblock Ethical and legal standards in social work: Consistency and conflict.
\newblock \emph{Families in society-The journal of contemporary social
  services}, 86\penalty0 (2):\penalty0 163--169, 2005.

\bibitem[Shen et~al.(2020)Shen, Qi, Li, Bi, and Wang]{shen2020hierarchical}
Shirong Shen, Guilin Qi, Zhen Li, Sheng Bi, and Lusheng Wang.
\newblock Hierarchical chinese legal event extraction via pedal attention
  mechanism.
\newblock In \emph{Proceedings of COLING}, pages 100--113, 2020.

\bibitem[Speicher et~al.(2018)Speicher, Heidari, Grgichlaca, Gummadi, Singla,
  Weller, and Zafar]{speicher2018a}
Till Speicher, Hoda Heidari, Nina Grgichlaca, Krishna~P Gummadi, Adish Singla,
  Adrian Weller, and Muhammad~Bilal Zafar.
\newblock A unified approach to quantifying algorithmic unfairness: Measuring
  individual \&group unfairness via inequality indices.
\newblock \emph{knowledge discovery and data mining}, pages 2239--2248, 2018.

\bibitem[Xiao et~al.(2018)Xiao, Zhong, Guo, Tu, Liu, Sun, Feng, Han, Hu, Wang,
  et~al.]{xiao2018cail2018}
Chaojun Xiao, Haoxi Zhong, Zhipeng Guo, Cunchao Tu, Zhiyuan Liu, Maosong Sun,
  Yansong Feng, Xianpei Han, Zhen Hu, Heng Wang, et~al.
\newblock Cail2018: A large-scale legal dataset for judgment prediction.
\newblock \emph{arXiv preprint arXiv:1807.02478}, 2018.

\bibitem[Xiao et~al.(2019)Xiao, Zhong, Guo, Tu, Liu, Sun, Zhang, Han, Hu, Wang,
  et~al.]{xiao2019cail2019-scm:}
Chaojun Xiao, Haoxi Zhong, Zhipeng Guo, Cunchao Tu, Zhiyuan Liu, Maosong Sun,
  Tianyang Zhang, Xianpei Han, Zhen Hu, Heng Wang, et~al.
\newblock Cail2019-scm: A dataset of similar case matching in legal domain.
\newblock \emph{arXiv: Computation and Language}, 2019.

\bibitem[Ye et~al.(2018)Ye, Jiang, Luo, and Chao]{ye2018interpretable}
Hai Ye, Xin Jiang, Zhunchen Luo, and Wenhan Chao.
\newblock Interpretable charge predictions for criminal cases: Learning to
  generate court views from fact descriptions.
\newblock In \emph{Proceedings of NAACL}, 2018.

\bibitem[Zafar et~al.(2017)Zafar, Valera, Rodriguez, and
  Gummadi]{zafar2017fairnessa}
Muhammad~Bilal Zafar, Isabel Valera, Manuel~Gomez Rodriguez, and Krishna~P
  Gummadi.
\newblock Fairness constraints: Mechanisms for fair classification.
\newblock In \emph{Proceedings of AISTATS}, 2017.

\bibitem[Zhong et~al.(2018)Zhong, Zhipeng, Tu, Xiao, Liu, and
  Sun]{zhong2018legal}
Haoxi Zhong, Guo Zhipeng, Cunchao Tu, Chaojun Xiao, Zhiyuan Liu, and Maosong
  Sun.
\newblock Legal judgment prediction via topological learning.
\newblock In \emph{Proceedings of EMNLP}, 2018.

\bibitem[Zhong et~al.(2020{\natexlab{a}})Zhong, Wang, Tu, Zhang, Liu, and
  Sun]{zhongiteratively}
Haoxi Zhong, Yuzhong Wang, Cunchao Tu, Tianyang Zhang, Zhiyuan Liu, and Maosong
  Sun.
\newblock Iteratively questioning and answering for interpretable legal
  judgment prediction.
\newblock In \emph{Proceedings of AAAI}, 2020{\natexlab{a}}.

\bibitem[Zhong et~al.(2020{\natexlab{b}})Zhong, Xiao, Tu, Zhang, Liu, and
  Sun]{zhong2020does}
Haoxi Zhong, Chaojun Xiao, Cunchao Tu, Tianyang Zhang, Zhiyuan Liu, and Maosong
  Sun.
\newblock How does nlp benefit legal system: A summary of legal artificial
  intelligence.
\newblock In \emph{Proceedings of ACL}, 2020{\natexlab{b}}.

\bibitem[Zhong et~al.(2020{\natexlab{c}})Zhong, Xiao, Tu, Zhang, Liu, and
  Sun]{zhong2020jec}
Haoxi Zhong, Chaojun Xiao, Cunchao Tu, Tianyang Zhang, Zhiyuan Liu, and Maosong
  Sun.
\newblock Jec-qa: A legal-domain question answering dataset.
\newblock In \emph{Proceedings of AAAI}, 2020{\natexlab{c}}.

\end{thebibliography}

\begin{appendix}

\section{Data Volume}

In this section, we show the data volume of each experiment.

Again, it is important to emphasize that we keep the amount of data uniform for each group in each experiment. Hence, the dataset sizes used for each experiment we report below are for each group.

In the main text, \textbf{Section 4.2} (Simulation Study) and \textbf{Section 4.4} (De-biasing Strategies Evaluation), for both experiments with theft and intentional injury, we control the size of the artificial dataset to 5000 per group. Besides, for the simulation study, we introduced cases with only one criminal charge in every single experiment. The entire experimental results are shown in Section~\ref{section:simulation_study_result}, and the results we report in the main text are for the crime of theft.

Table~\ref{table:statistics} shows the dataset sizes used for the different crime experiments in \textbf{Section 4.3} (\metricname{} on Realistic Dataset) of the main text. In addition to this, for experiments using data from one specific region (i.e., the experiments in Table 3 of the main text), the amount of data for the individual charge experiments is also the same as that presented in Table~\ref{table:statistics}.

\begin{table}[ht]
        \centering
        \small
        \tabcolsep 30pt
        \caption{Statistics for the data set of Table 2, and Table 3 experiment in the main text.}
    \label{table:statistics}
    \begin{tabular*}{\textwidth}{ccc}
    \toprule
        \textbf{Crime} & \textbf{Data per Region} & \textbf{Data per Gender}\\ \midrule
        Fraud & 2000 & 5000 \\ 
        Drug Trafficking & 2000 & 5000 \\ 
        \tabincell{c}{Possession of Illegal Drugs} & 500 & 3031 \\ 
        Intentional Injury & 5000 & 5000 \\ 
        Traffic Offence & 2000 & 5000 \\ 
        Theft & 5000 & 5000\\ 
        \tabincell{c}{Picking Quarrels and Provoking Trouble} & 1000 & 1740 \\ 
        \tabincell{c}{Disrupting Public Service} & 500 & 5000 \\ 
        \tabincell{c}{Providing Venues for Drug Users} & 500 & 5000 \\
    \bottomrule
    \end{tabular*}
    
\end{table}

Also, in \textbf{Section 4.3}, the amount of data used for each count in experiments with regional inconsistencies across different years (i.e., the experiments in Table 4 of the main text) is shown in Table~\ref{table:statistics_different_year}.

\begin{table}[ht]
        \centering
        \small
        \tabcolsep 35pt
        \caption{Statistics for the data set of the Table 5 experiment in the main text.}
    \label{table:statistics_different_year}
    \begin{tabular*}{\textwidth}{ccc}
    \toprule
        \textbf{Crime} & \textbf{2013 -- 2015} & \textbf{2016 -- 2018} \\\midrule
        Fraud & 1000 & 1000\\ 
        Drug Trafficking & 1000 & 1000\\ 
        Intentional Injury & 4800 & 4800\\ 
        Traffic Offence & 1000 & 1000\\ 
        Theft & 5000 & 5000\\ 
        \tabincell{c}{Picking Quarrels and Provoking Trouble} & 500 & 500\\
    \bottomrule
    \end{tabular*}
    
\end{table}

\section{Experimental Results of The Simulation Study}
\label{section:simulation_study_result}

The entire experimental results of the simulation study are displayed in Table~\ref{table:copy}. Note that the results we report in the main text are for the crime of theft.

\begin{table*}[t]
    \centering
\caption{
    The result of different $\beta$ and $K$. For each cell, we conduct multiple experiments and report the mean and standard deviation of \metricname{}.
    }
    \label{table:copy}
\resizebox{1.0\textwidth}{!}{
    \begin{tabular}{c|c|c|ccccccc}\toprule
         \multicolumn{3}{c|}{$\beta$} & $0$ & $0.1$ & $0.2$ & $0.3$ & $0.4$ & $0.5$ & $1.0$ \\ \midrule

 &  & $K=2$ & $0.200\pm0.028$  & $0.214\pm0.038$  & $0.237\pm0.018$  & $0.288\pm0.026$  & $0.337\pm0.053$  & $0.462\pm0.071$  & $1.294\pm0.276$ \\
 &  & $K=3$ & $0.262\pm0.037$  & $0.269\pm0.028$  & $0.290\pm0.046$  & $0.315\pm0.050$  & $0.366\pm0.029$  & $0.431\pm0.074$  & $0.891\pm0.108$ \\
 & \metricname{} & $K=4$ & $0.283\pm0.030$  & $0.306\pm0.036$  & $0.310\pm0.040$  & $0.331\pm0.040$  & $0.388\pm0.064$  & $0.427\pm0.064$  & $0.960\pm0.156$ \\
 & with DGN & $K=5$ & $0.301\pm0.029$ & $0.303\pm0.029$  & $0.329\pm0.027$  & $0.354\pm0.026$  & $0.384\pm0.054$  & $0.438\pm0.043$  & $0.944\pm0.130$ \\
 &  & $K=6$ & $0.302\pm0.019$  & $0.316\pm0.029$  & $0.335\pm0.040$  & $0.370\pm0.025$  & $0.404\pm0.068$  & $0.462\pm0.035$  & $0.793\pm0.130$ \\
 &  & $K=7$ & $0.325\pm0.031$  & $0.325\pm0.026$  & $0.334\pm0.031$  & $0.372\pm0.044$  & $0.399\pm0.040$  & $0.461\pm0.046$  & $0.842\pm0.104$ \\
\cline{2-10} &  & $K=2$ & $0.000\pm0.000$ & $0.079\pm0.008$ & $0.166\pm0.017$ & $0.257\pm0.028$ & $0.362\pm0.031$ & $0.500\pm0.074$ & $1.437\pm0.174$ \\
Theft &  & $K=3$ & $0.000\pm0.000$ & $0.066\pm0.007$ & $0.134\pm0.014$ & $0.206\pm0.024$ & $0.289\pm0.027$ & $0.390\pm0.034$ & $1.024\pm0.107$ \\
 & Golden & $K=4$ & $0.000\pm0.000$ & $0.060\pm0.007$ & $0.122\pm0.012$ & $0.189\pm0.020$ & $0.261\pm0.029$ & $0.341\pm0.040$ & $0.906\pm0.162$ \\
 & Value & $K=5$ & $0.000\pm0.000$ & $0.057\pm0.006$ & $0.116\pm0.012$ & $0.181\pm0.019$ & $0.250\pm0.025$ & $0.324\pm0.032$ & $0.814\pm0.094$ \\
 &  & $K=6$ & $0.000\pm0.000$ & $0.056\pm0.006$ & $0.113\pm0.010$ & $0.174\pm0.018$ & $0.239\pm0.022$ & $0.312\pm0.032$ & $0.788\pm0.057$ \\
 &  & $K=7$ & $0.000\pm0.000$ & $0.054\pm0.006$ & $0.110\pm0.011$ & $0.168\pm0.018$ & $0.233\pm0.020$ & $0.304\pm0.035$ & $0.752\pm0.102$ \\
 &  & $K=100$ & $0.000\pm0.000$ & $0.049\pm0.005$ & $0.098\pm0.010$ & $0.149\pm0.016$ & $0.204\pm0.021$ & $0.261\pm0.027$ & $0.638\pm0.058$ \\
\midrule &  & $K=2$ & $0.153\pm0.025$  & $0.169\pm0.049$  & $0.183\pm0.034$  & $0.252\pm0.042$  & $0.371\pm0.053$  & $0.429\pm0.056$  & $1.260\pm0.169$ \\
 &  & $K=3$ & $0.186\pm0.036$  & $0.179\pm0.018$  & $0.212\pm0.023$  & $0.276\pm0.040$  & $0.307\pm0.044$  & $0.384\pm0.050$  & $0.934\pm0.079$ \\
 & \metricname{} & $K=4$ & $0.190\pm0.031$  & $0.195\pm0.030$  & $0.230\pm0.039$  & $0.267\pm0.049$  & $0.313\pm0.033$  & $0.367\pm0.038$  & $0.771\pm0.095$ \\
 & with DGN & $K=5$ & $0.230\pm0.038$  & $0.250\pm0.055$  & $0.249\pm0.046$  & $0.272\pm0.021$  & $0.317\pm0.050$  & $0.379\pm0.066$  & $0.760\pm0.090$ \\
 &  & $K=6$ & $0.223\pm0.025$  & $0.224\pm0.027$  & $0.246\pm0.031$  & $0.264\pm0.041$  & $0.304\pm0.033$  & $0.391\pm0.037$  & $0.786\pm0.134$ \\
 &  & $K=7$ & $0.228\pm0.037$  & $0.229\pm0.028$  & $0.246\pm0.033$  & $0.270\pm0.045$  & $0.322\pm0.026$  & $0.371\pm0.045$  & $0.821\pm0.133$ \\
\cline{2-10} Intentional &  & $K=2$ & $0.000\pm0.000$ & $0.074\pm0.007$ & $0.152\pm0.015$ & $0.244\pm0.021$ & $0.336\pm0.036$ & $0.469\pm0.058$ & $1.502\pm0.208$ \\
 Injury &  & $K=3$ & $0.000\pm0.000$ & $0.060\pm0.006$ & $0.123\pm0.011$ & $0.192\pm0.018$ & $0.270\pm0.026$ & $0.358\pm0.042$ & $0.885\pm0.089$ \\
 & Golden & $K=4$ & $0.000\pm0.000$ & $0.056\pm0.005$ & $0.114\pm0.010$ & $0.176\pm0.016$ & $0.239\pm0.020$ & $0.314\pm0.036$ & $0.836\pm0.088$ \\
 & Value & $K=5$ & $0.000\pm0.000$ & $0.053\pm0.005$ & $0.109\pm0.010$ & $0.165\pm0.015$ & $0.229\pm0.020$ & $0.304\pm0.027$ & $0.775\pm0.086$ \\
 &  & $K=6$ & $0.000\pm0.000$ & $0.052\pm0.005$ & $0.104\pm0.008$ & $0.160\pm0.015$ & $0.222\pm0.021$ & $0.287\pm0.026$ & $0.734\pm0.060$ \\
 &  & $K=7$ & $0.000\pm0.000$ & $0.051\pm0.004$ & $0.102\pm0.009$ & $0.156\pm0.014$ & $0.216\pm0.018$ & $0.278\pm0.026$ & $0.719\pm0.049$ \\
 &  & $K=100$ & $0.000\pm0.000$ & $0.045\pm0.004$ & $0.091\pm0.008$ & $0.139\pm0.012$ & $0.190\pm0.017$ & $0.243\pm0.022$ & $0.594\pm0.049$ \\

\bottomrule
    \end{tabular}
}
\end{table*}

\section{Description of charges}

In this section, we describe the criminal charges mentioned in this paper. All the descriptions refer to Chinese Criminal Law.

\textbf{Fraud}: The act that, for the purpose of illegal possession, swindles public or private money or property, and if the amount is relatively large.

\textbf{Drug Trafficking}: The act of knowingly smuggling, trafficking, transporting or manufacturing drugs.

\textbf{Possession of Illegal Drugs}: The act that illegally possesses drugs (e.g., opium, heroin, methylaniline) knowingly of relatively large quantities.

\textbf{Intentional Injury}: The act that intentionally inflicts injury upon another person.

\textbf{Traffic Offence}: The act violates regulations governing traffic and transportation and thereby causes a serious accident, resulting in serious injuries or deaths or heavy losses of public or private property.

\textbf{Theft}: the act that, for the purpose of illegal possession, steals a relatively large amount of public or private property or commits theft repeatedly.

\textbf{Picking Quarrels and Provoking Trouble}: Committing any of the following acts of creating disturbances, thus disrupting public order: (1) beating another person at will and to a flagrant extent; (2) chasing, intercepting or hurling insults to another person to a flagrant extent; (3) forcibly taking or demanding, willfully damaging, destroying or occupying public or private money or property to a serious extent; or (4) creating disturbances in a public place, thus causing serious disorder in such place.

\textbf{Disrupting Public Service}: The act that, by means of violence or threat, obstructs a functionary of a State organ from carrying out his functions according to law.

\textbf{Providing Venues for Drug Users}: The act that provides shelter for another person to ingest or inject narcotic drugs.

\end{appendix}

\end{document}